\documentclass[10pt,twocolumn,oneside,letterpaper]{article} 

\usepackage{avss}
\usepackage{times}
\usepackage{epsfig}
\usepackage{graphicx}
\usepackage{amsmath}
\usepackage{amssymb}
\usepackage{booktabs}
\usepackage{multicol}

\usepackage{algorithm,algpseudocode}
\usepackage{comment}
\usepackage{adjustbox}
\usepackage{multirow}
\usepackage{float}
\usepackage{pifont}
\usepackage{xcolor}
\usepackage{fancyhdr,lipsum}

\usepackage[pagebackref=true,breaklinks=true,letterpaper=true,colorlinks,bookmarks=false]{hyperref}

\avssfinalcopy % *** Uncomment this line for the final submission

\def\avssPaperID{77} % *** Enter the AVSS Paper ID here

\ifavssfinal\fi
\begin{document}

%%%%%%%%% TITLE

\title{Spatio-temporal predictive tasks for abnormal event detection in videos}

\author{{ {Yassine Naji$^{1,2}$, Aleksandr Setkov$^{1}$, Angélique Loesch$^{1}$, Michèle Gouiffès$^{2}$, Romaric Audigier$^{1}$}}\\
{\normalsize $^{1}$Université Paris-Saclay, CEA, List, 91120, Palaiseau, France} \\  
{\tt\small firstname.lastname@cea.fr}
\\
{\normalsize $^{2}$ Université Paris-Saclay, CNRS, LISN, 91400, Orsay, France}
\\
{\tt\small firstname.lastname@universite-paris-saclay.fr}
}

\maketitle
\thispagestyle{empty}

\newcommand{\blue}[1]{\textcolor{black}{#1}}
\newcommand{\green}[1]{\textcolor{black}{#1}}

%%%%%%%%% ABSTRACT A
\begin{abstract}
  Abnormal event detection in videos is a challenging problem, partly due to the multiplicity of abnormal patterns and the lack of their corresponding annotations. In this paper, we propose new constrained pretext tasks to learn object level normality patterns. Our approach consists in learning a mapping between down-scaled visual queries and their corresponding normal appearance and motion characteristics at the original resolution. The proposed tasks are more challenging than reconstruction and future frame prediction tasks which are widely used in the literature, since our model learns to jointly predict spatial and temporal features rather than reconstructing them. We believe that more constrained pretext tasks induce a better learning of normality patterns. Experiments on several benchmark datasets demonstrate the effectiveness of our approach to localize and track anomalies as it outperforms or reaches the current state-of-the-art on spatio-temporal evaluation metrics.
\end{abstract}

\let\thefootnote\relax\footnote{978-1-6654-6382-9/22/\$31.00 ©2022 IEEE}

%%%%%%%%% BODY TEXT
\section{Introduction}

\begin{figure}[!t]
    \centering
    \includegraphics[width=0.42\textwidth]{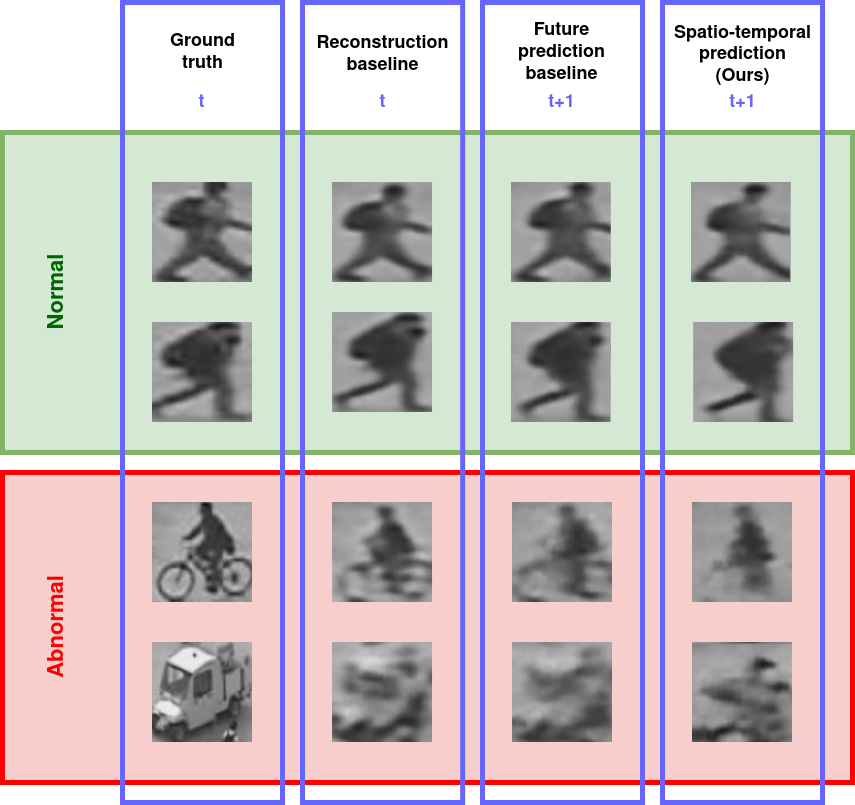}
    \label{fig:teaser}
    \caption{Normal and abnormal test samples with their corresponding predictions using 3 different anomaly detection pretext tasks: \textit{spatio-temporal prediction} (ours), \textit{future frame prediction} and spatial \textit{reconstruction}. The use of spatio-temporal prediction tasks leads to a good recovery of normal objects and a worse recovery of abnormal samples compared to other pretext tasks.}
\end{figure} 

Video anomaly detection (VAD) is an open research problem which consists in detecting the occurrence of unexpected events. This problem has raised a lot of attention due to its critical applications such as video surveillance \cite{ucfcrime,shanghaitech,ped,avenue,streetscene,adoc} and autonomous driving \cite{dada}. For several reasons, this is a challenging problem. \blue{First, an event is considered as abnormal according to a set of normal events which define a context}, therefore, the same event can be considered normal or abnormal in two different contexts. Second, the wide range of possible abnormal events and their rarity make it infeasible to collect enough annotations to train fully supervised models. Consequently, VAD is often regarded as a "one class" problem, when only normal data is used for training. \blue{ Several approaches have been proposed in the literature to address this challenge. They can be grouped into three main paradigms: \textbf{Distance based methods} \cite{dummyae,streetscene,seamese,clusters_ionescu,plug_and_play,local_stats} 
which are trained by learning a distance between samples. At inference time, the anomaly score is computed as the distance of a sample to normal data. \textbf{Probabilistic methods} \cite{survey_ramachandra} learn a distribution over normal data. Samples with low likelihood according to the estimated probability density are detected as anomalies. \textbf{Self-supervised methods} \cite{baf,mnad,memae,app-mo-co,robust,temp_reg,sparse,avenue,stacked_rnn, future_pred,ssmt,ganicip17,discrete,hybrid,dual_discriminator,Tang} use some pretext tasks to learn normal appearance and motion features from training data. At inference time, anomalies are often detected by measuring the incapacity of the model to perform a pretext task on a test sample. The choice of the pretext task conditions the type of anomalies the model is able to detect. Therefore, these tasks must be chosen according to the anomalies of interest.}

A widely used self-supervised approach for anomaly detection consists in \textit{reconstructing} normal samples from a low dimensional representations \cite{baf,mnad,memae,app-mo-co,robust,temp_reg,sparse,avenue,stacked_rnn}. The underlying assumption is that a model, trained on normal data only, can not generalize well to abnormal samples. Other approaches \cite{future_pred,ssmt,ganicip17,discrete,hybrid,dual_discriminator,Tang} use \textit{prediction} based pretext tasks to learn normal features from training data. Those approaches achieve good performances, however, \blue{we believe that it is important to further constrain those tasks to better learn normality patterns and thus better discriminate anomalies} (figure \ref{fig:teaser}). For instance, a common approach consists in training a neural network to predict the next frame from a series of past frames \cite{future_pred}. Given a video sequence of a person walking in a wrong direction, \blue{the model could infer the location of the person in the next frame by referring to the past}. Even if this event is abnormal, it is potentially predictable if the model learns the concept of walking from normal data. As an alternative, we propose to discard past frames and to constrain the model to infer both the next and past frames \blue{\textit{using only the current frame}} by generalizing from training data and not by referring to past frames. The task is constrained further by \textit{down-scaling} the input in order to recover as least as possible abnormal features. \blue{Thus, the network is forced to predict spatial features in addition to temporal ones}. More precisely, we train our model to predict, in the original resolution, the appearance and optical flow of a given object while it is conditioned only on a down-scaled version of it. At test phase, abnormal samples are detected as those that produce prediction errors which are significantly higher than the normal training samples. We measure this discrepancy using a \textit{Z-score} based normalization. In summary our contributions are:

$\bullet$ Introducing new constrained pretext tasks for anomaly detection \blue{based on spatio-temporal normality prediction. Specifically, by predicting jointly upscaled version of object appearance and motion}.
 
$\bullet$ Showing empirically that our method outperforms or reaches the current state-of-the-art on spatio-temporal evaluation metrics.

Since our method is designed for object level anomalies, we tested it on the three common benchmark datasets: UCSDped2 \cite{ped}, ShanghaiTech \cite{shanghaitech} and Avenue \cite{avenue} for which the anomalies are related to object appearance and behaviour. The experiments conducted on those datasets show the relevance of the proposed approach.

\section{Related works}

%\pl{Reconstruction prediction pretext task}
\paragraph*{Anomaly detection pretext tasks:} Recently, deep learning approaches \cite{baf,ssmt,dummyae,hybrid,future_pred,memae,mnad,ganicip17,streetscene,seamese,dual_discriminator,cloze_test,Tang,app-mo-co,clusters_ionescu} have shown their effectiveness for detecting abnormal events in videos. Those approaches learn normality using some pretext tasks. Among those tasks, we can distinguish two major families: \textit{reconstruction} and \textit{prediction} tasks.

Many previous methods used \textit{reconstruction} to learn normality \cite{baf,memae,temp_reg,mnad,app-mo-co,robust,sparse,avenue}. In a pioneering work \cite{temp_reg}, the authors proposed to train an auto-encoder to reconstruct normal handcrafted appearance and motion features. The reconstruction task is constrained further by adding a memory module which learns normality prototypes \cite{memae}. An extension of this approach was proposed by \cite{mnad} which learns spatio-temporal patch prototypes. 

%\pl{Futur frame prediction pretext task}

Other methods such as \cite{future_pred,dual_discriminator,discrete,ganicip17,app-mo-co,discrete,Tang,background_sub} trained GANs to perform \textit{prediction} tasks. A common way consists in training a generator to predict future frame \blue{or equivalently the optical flow} given few past frames \cite{future_pred,dual_discriminator,Tang}. Abnormal events are detected when the prediction error is high. 

%This work has been extended  \cite{dual_discriminator} by adding an optical flow discriminator to focus on more the motion information. 

Similarly to adversarial approaches, we learn to predict normal appearance and motion patterns. Unlike those approaches, we do not use a discriminator during training since we observed that our method is already able to produce consistent predictions using our objective function. 

\blue{Approaches \cite{future_pred,dual_discriminator,Tang} performed prediction at the temporal level. We propose to further constrain the normality learning task by imposing prediction at both the spatial and temporal levels. Our model learns up-scaling in conjunction with future and past prediction.} Moreover, the normality is learned at the object level to ensure the invariance to scene changes, unlike previous works \cite{future_pred,dual_discriminator,Tang} which model normality at the frame level and therefore they are sensitive to background changes.

\paragraph*{Object centric anomaly detection:} Some recent works characterized abnormality at the object level \cite{ssmt,baf,dummyae,hybrid,any_shot,cloze_test}, by making use of a pretrained object detector as a preliminary step before anomaly scoring. Ionescu \textit{et al.} \cite{dummyae} proposed to learn appearance and motion features using an auto-encoder and performed a clustering followed by a one versus rest classification. Recently, Georgescu \textit{\textit{et al.}} \cite{baf} introduced a new approach that uses out-of-domain observations to train a model to explicitly mis-reconstruct those pseudo-anomalies via adversarial training. \blue{This approach achieves very good performance on current benchmark datasets, nevertheless, it needs additional pseudo-abnormal data for training}. We propose to train our model using normal data only without explicit assumptions on anomalies, which makes our method generalizable to multiple anomaly types.

 Liu \textit{\textit{et al.}} \cite{hybrid} extended \cite{future_pred} by applying the future frame prediction task at the object level with a constraint on the reconstruction of optical flows via a multi-level memory module. The addition of this module restricts the model ability to recover anomalies, however, it requires choosing the number of memory items which models the heterogeneity of normality. We propose not to discretize the normality space. Instead, we constrain anomaly prediction by discarding past frames and inputting only a down-scaling of the current frame. In addition, we perform optical flow prediction instead of reconstruction so that the model cannot reproduce abnormal motion features.

By observing that single objective optimization is generally suboptimal for anomaly detection, Georgescu \textit{\textit{et al.}} proposed in \cite{ssmt} a multitask architecture to learn normality through multiple pretext tasks such as middle frame prediction and arrow of time prediction. \blue{We propose to constrain further the pretext tasks to better learn normality patterns} \ref{fig:teaser}.  Specifically, our model learns to restore normal appearance and motion from a down-scaled image of the object. To our knowledge, we are the first to propose \textit{super-resolution} as a pretext task for video anomaly detection.

\section{Method}

\subsection{Motivation}

The main motivation behind our approach is the following: \textit{self-supervised pretext tasks for which the target cannot be predicted directly from the input but only by generalizing from the training data manifold are more suited for anomaly detection}.

\begin{comment}
We believe that the model should infer the normal version of the output only by generalizing from training data and relying as less as possible on the input, which should serves only as a query for the corresponding normality. We hypothesis that those types of pretext tasks are better at discriminating between abnormal and normal samples. 
\end{comment}

We believe that if the model do not access the necessary information to predict the target from the input, it is constrained to generalize from training data to perform the task. Since we assume that the model is trained on normal data only, it will induce a larger prediction error for abnormal samples. Motivated by this intuition, we propose to further constrain pretext tasks by training the model to perform future/past frame prediction using only a down-scaling of the current frame, which is more challenging than future frame prediction proposed in \cite{future_pred,hybrid,dual_discriminator,Tang} where past frames are given as input to perform the task. \blue{In addition to the common temporal prediction \cite{future_pred,hybrid,dual_discriminator,Tang}, our model performs spatial prediction by up-scaling the input}. Specifically, we train our method via four pretext tasks which can be grouped as follows: 

\textbf{\textit{Appearance tasks}}: consist in predicting the original resolution of an object image in both past and next frames.

\textbf{\textit{Motion tasks}}: consist in predicting the forward optical flow magnitudes at both the past and the current frames. 

\begin{comment}

\end{comment}
\begin{figure}[t]
    \centering
    \begin{minipage}[b]{0.45\textwidth}
        \includegraphics[width=1\textwidth,trim={3cm 2cm 2cm 1cm},clip]{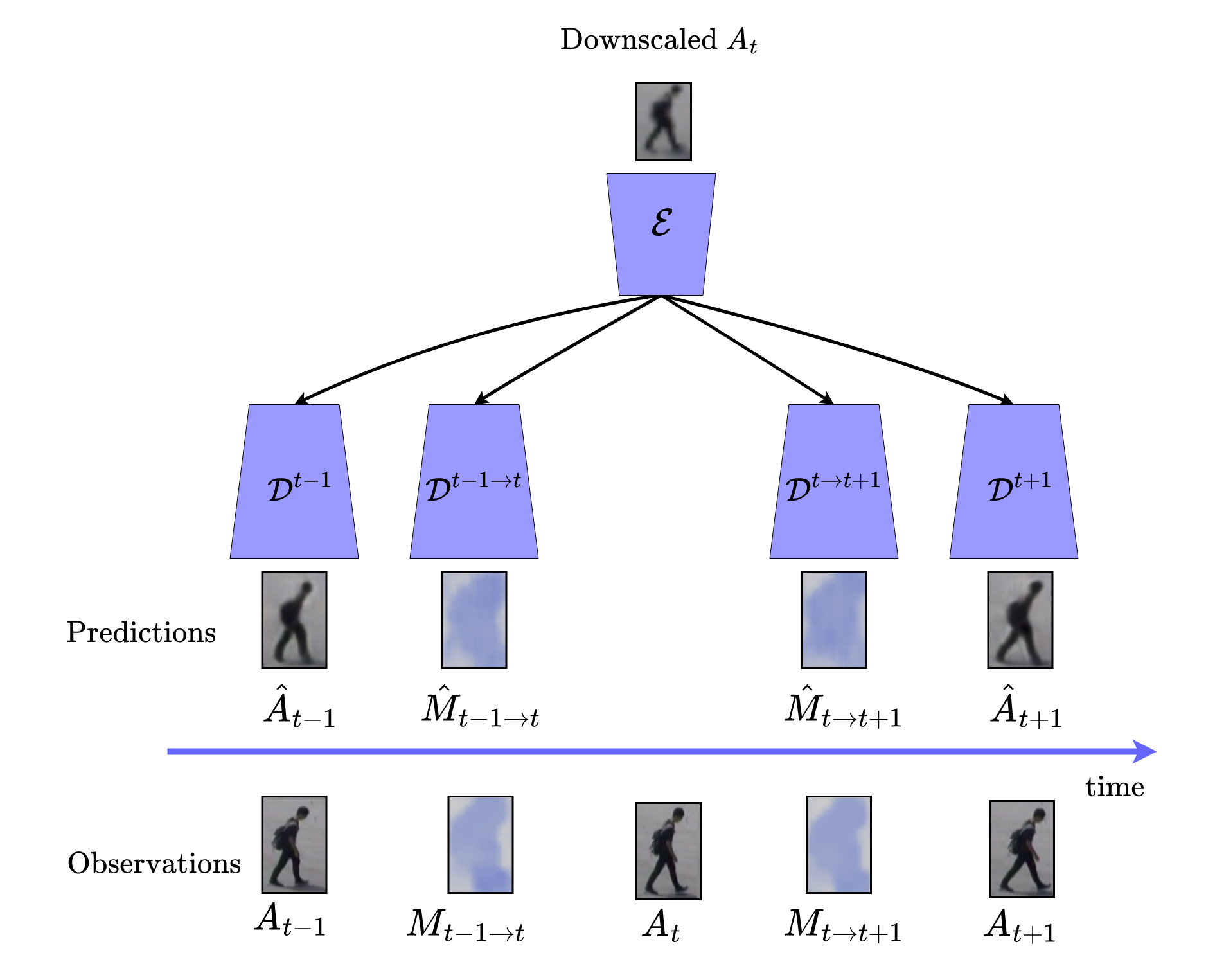}
        \label{fig:architecture}

    \end{minipage}
    \caption{Overview of our architecture: Considering a frame $t$, we perform object detection and forward optical flow extraction: $t-1  \rightarrow t$ and $t \rightarrow t+1$. The object appearances at frames $t - 1$, $t$ and $t + 1$: $A_{t-1} , A_{t}, A_{t+1}$ with its corresponding optical flow magnitudes: $M_{t-1 \rightarrow t} , M_{t \rightarrow t+1} $ are cropped. The object appearance at time step $t$: $A_t$ is down-scaled and is then passed as an input to the encoder $\mathcal{E}$ that produces a latent representation which is given to the four decoding branches $\mathcal{D}$.  Appearances:  $ \hat{A}_{t-1},\hat{A}_{t+1}$ and optical flow magnitudes $\hat{M}_{t-1 \rightarrow t} , \hat{M}_{t \rightarrow t+1} $ are predicted by those branches \blue{at the original resolution}}
\end{figure}

\begin{comment}

\begin{figure*}
	\centering
	\includegraphics[width=0.6\textwidth]{figs/schema_avss.drawio.png}
	\caption{Overview of our pipeline: Considering a time step $t$, we perform object detection and forward optical flow extraction: $t-1  \rightarrow t$ and $t \rightarrow t+1$. The object at time step $t - 1$, $t$ and $t + 1$ with its corresponding optical flow magnitudes are cropped. The object at time step $t$ is down-scaled and is given as an input to the encoder that produces a latent representation which is given to the four decoding branches.  Appearances and optical flow magnitudes are predicted by those branches.}
	\label{fig:architecture}
\end{figure*}
\end{comment}

\subsection{Approach}

\subsubsection{Inputs and pre-processing}

In order to localize anomalies at the object level, we first perform object detection similarly to other object-centric approaches \cite{ssmt,baf,dummyae}. We crop the detected objects using bounding boxes at frames $F_{t - 1}$, $F_{t}$ and $F_{t + 1}$. 
We do not perform tracking since we want to capture the absolute object displacement. We suppose that the position change between times $t-1, t+1$ is small enough so that the majority of the object is still contained in the bounding box. \blue{In order to constrain further the prediction tasks, we propose to down-scale} objects at frame $F_t$, which are fed as input to our architecture. 

The forward optical flow is computed between the frames $F_t$ and $F_{t+1}$ and the magnitude map is extracted. We do not keep the orientation map since it is not precise enough for small magnitude displacements. The magnitude map is cropped using the bounding boxes and resized to the same objects resolution.

\subsubsection{Architecture}
Since our model performs multiple pretext tasks, we choose an architecture with multiple predictive branches. More precisely, it is composed of a common encoder $\mathcal{E}$ \green{which learns high level appearance and motion features} and four decoding branches. $\mathcal{D}^{t-1}$, $\mathcal{D}^{t-1 \rightarrow t}$ are dedicated to predict past context and the two other branches $\mathcal{D}^{t\rightarrow t+1}, \mathcal{D}^{t+1} $ are dedicated to future context. In particular, the model learns to recover \blue{from low-resolution input}, normal appearance and motion jointly via predicting the original resolution of the object in both previous and next frames using the branches $\mathcal{D}^{t-1}$ and $\mathcal{D}^{t+1}$. In order to emphasize learning motion patterns, we train the model to predict the magnitude of optical flow via the branches $\mathcal{D}^{t-1 \rightarrow t}$ and $\mathcal{D}^{t \rightarrow t+1}$. The architecture takes as input a down-scaled object image and outputs past and future object appearances: $\hat{A}_{t-1}$ and $\hat{A}_{t+1}$ with the corresponding optical flow magnitudes $\hat{M}_{t-1\rightarrow t}$ and $\hat{M}_{t \rightarrow t+1}$. The figure \ref{fig:architecture} illustrates our network.

\subsubsection{Loss functions}
\begin{comment}

\begin{equation}
    \mathcal{L}_{context} =  \mathcal{L}_{past} + \lambda \mathcal{L}_{future} 
\end{equation}

\textit{Past context loss}: composed of an appearance term which is a logistic loss between the prediction of the branch the object appearance $O^{t-1}$ predicted by the branch $\mathcal{D}^{t-1}$ and the ground-truth object image. We use the same loss between the predicted optical flow magnitude $\hat{M}_{t-1 \rightarrow t}$ at branch $\mathcal{D}^{t-1\rightarrow t}$ and the ground-truth $M_{t-1 \rightarrow t}$: 

\begin{equation}
    \mathcal{L}_{past} =   \mathcal{L}(\hat{A}_{t-1},A_{t-1}) +\mathcal{L}(\hat{M}_{t-1 \rightarrow t},M_{t-1 \rightarrow t})
\end{equation}

Where: 
{\small
\begin{equation}
    \mathcal{L}(\hat{X},X) = \sum_{i,j } - X_{i,j} \log (\hat{X}_{i,j}) - (1-X_{i,j}) \log(1-\hat{X}_{i,j})
\end{equation}
}

\textit{Future context loss}: similarly to $\mathcal{L}_{past}$, we define:
\begin{equation}
    \mathcal{L}_{future} =   \mathcal{L}(\hat{A}_{t+1},A_{t+1}) +\mathcal{L}(\hat{M}_{t\rightarrow t+1},M_{t \rightarrow t+1})
\end{equation}

\end{comment}

In order to learn normal appearance and motion patterns, we train our framework to predict the normal temporal context using a combination of two prediction losses: 
%\pl{peut etre ne pas introduire les $\lambda$ montre qu'il n'y a pas d'hyper parametres à tuner}
\begin{equation}
    \mathcal{L}_{context} =  \mathcal{L}_{past} + \mathcal{L}_{future}
\end{equation}
with:
\begin{equation}
    \mathcal{L}_{past} =   \mathcal{L}(\hat{A}_{t-1},A_{t-1}) +  \mathcal{L}(\hat{M}_{t-1 \rightarrow t},M_{t-1 \rightarrow t})
\end{equation}
\begin{equation}
     \mathcal{L}_{future} = \mathcal{L}(\hat{A}_{t+1},A_{t+1}) + \mathcal{L}(\hat{M}_{t\rightarrow t+1},M_{t \rightarrow t+1})
\end{equation}

where $\mathcal{L}$ is the logistic loss computed over all pixel locations $i$, $j$:

\begin{equation}
    \mathcal{L}(\hat{X},X) = \sum_{i,j } - X_{i,j} \log (\hat{X}_{i,j}) - (1-X_{i,j}) \log(1-\hat{X}_{i,j})
\end{equation}

The past context loss $\mathcal{L}_{past}$ is composed of an appearance term which is a logistic loss between the prediction of the branch $\mathcal{D}^{t-1}$ prediction $\hat{A}_{t-1}$ and ${A}_{t-1}$. We use the same loss between the predicted optical flow magnitude $\hat{M}_{t-1 \rightarrow t}$ at branch $\mathcal{D}^{t-1\rightarrow t}$ and the observation $M_{t-1 \rightarrow t}$.

The future context loss $\mathcal{L}_{future}$ is defined in a similar way, between times $t$ and $t+1$.

\subsubsection{Inference: anomaly scoring} \label{sec:scoring}

At inference time, in order to compute the object level anomaly scores given a test video sequence, we perform object detection with the same parameters as for training. Similarly to training data, we also extract the frame level optical flow magnitudes and we crop them at the object level using the predicted boxes. For each test sample $O_t$, we compute the $\ell_{1}$ prediction errors for appearance and motion: 

\begin{comment}
\begin{equation}
    L^{t-1}(O_t) = \|\hat{A}_{t-1} - A_{t-1}\|_{1} 
\end{equation}

\begin{equation}
    L^{t-1 \rightarrow t}(O_t) = \|\hat{M}_{t-1 \rightarrow t} - M_{t-1 \rightarrow t}\|_{1} 
\end{equation}

\begin{equation}
    L^{t \rightarrow t+1}(O_t) =\|\hat{M}_{t \rightarrow t+1} - M_{t \rightarrow t+1}\|_{1}
\end{equation}

\begin{equation}
    L^{t+1}(O_t) = \|\hat{A}_{t+1} - A_{t+1}\|_{1} 
\end{equation}

Resulting on a 4D vector: 

\begin{equation}
    L(O_t) = \begin{pmatrix}
    L^{t-1}(O_t)\\
    L^{t-1 \rightarrow t}(O_t)  \\
    L^{t \rightarrow t+1}(O_t) \\
    L^{t+1}(O_t)
    \end{pmatrix}
\end{equation}
\end{comment}

\begin{equation}
    L(O_t) = \begin{pmatrix}
    L^{t-1}(O_t)\\
    L^{t-1 \rightarrow t}(O_t)  \\
    L^{t \rightarrow t+1}(O_t) \\
    L^{t+1}(O_t)
    \end{pmatrix} = \begin{pmatrix}
 \|\hat{A}_{t-1} - A_{t-1}\|_{1}    \\
 \|\hat{M}_{t-1 \rightarrow t} - M_{t-1 \rightarrow t}\|_{1} \\
 \|\hat{M}_{t \rightarrow t+1} - M_{t \rightarrow t+1}\|_{1}\\
\|\hat{A}_{t+1} - A_{t+1}\|_{1} \\
    \end{pmatrix}
\end{equation}

In order to compare this vector with respect to the distribution of normal samples, we propose to compute the Z-scores across all branches and average them. More precisely, we estimate the mean $\mu$ and the covariance matrix $\Sigma$ of $L$ vectors of normal training objects.

If we denote $n = 4$, the dimension of the vector $L$, the aggregated score is given by: 
\begin{equation}
    S(O_t) = \frac{1}{n}\text{diag}(\Sigma)^{-\frac{1}{2}} (L(O_t) - \mu)
\end{equation}

\noindent Let $\Delta_t$ be the set of $N_t$ detected objects at frame $F_t$: 
\begin{equation*}
    \Delta_t = \{ O_t^{(i)}, \quad 1 \leq i \leq N_t\}
\end{equation*}

\noindent Since we consider a frame $F_t$ abnormal if it contains at least one abnormal object, the frame level anomaly score is given by: 

 \begin{equation}
    S(F_t) =\max_{O \in \Delta_t}  S(O)
\end{equation}

\section{Experiments and results}
\label{sec:exp_res}

\subsection{Datasets}

Several benchmarks exist for evaluating anomaly detection performances \cite{ped,avenue,shanghaitech,streetscene,ucfcrime}. In order to compare our method with previous works, we performed experiments on the most common datasets.

\textbf{UCSDped2} \cite{ped} contains 16 training and 12 testing videos of resolution 240x360. Anomalous events include riding a bike and driving a vehicle on a sidewalk. The original dataset is annotated at both the frame and the pixel level. Ramachandra \textit{\textit{et al.}} \cite{streetscene} provided region-level and track-level annotations.

\textbf{ShanghaiTech} \cite{shanghaitech} includes 330 training and 107 testing videos with the resolution of 480x856 with 13 different scenes. Each scene has a different background. Abnormal events include jumping, running, or stalking on a sidewalk. As UCSDped2, the original dataset is annotated at frame and pixel levels. The region-level and track-level annotations are provided by Georgescu et al \cite{baf}.

\textbf{CUHK Avenue} \cite{avenue} consists of 16 training and 21 test videos with the resolution of 360x640  with abnormal events such as running or walking towards the camera. We evaluated our method on all metrics using the ground truth region-level and track-level annotations provided by Ramachandra \textit{\textit{et al.}} \cite{streetscene} for this dataset. Unlike other datasets, some annotated anomalies are related to the position with respect to the camera. Therefore, we take into account object sizes for measuring abnormality as explained in section \ref{sec:avenue_res}.
 
\begin{table*}[!t]
        \centering
        \resizebox{0.9\linewidth}{!}{%
		\begin{tabular}{llcccccccccccccc}
%		    \cmidrule(lr){2-14}
		    \toprule
		    &\multirow{3}{*}{Method}&\multicolumn{4}{c}{UCSDped2}&\multicolumn{4}{c}{ShanghaiTech}&\multicolumn{4}{c}{Avenue} \\
			 \cmidrule(lr){3-14}
			 &&\multicolumn{2}{c}{AUC}&\multirow{2}{*}{RBDC}&\multirow{2}{*}{TBDC}&\multicolumn{2}{c}{AUC}&\multirow{2}{*}{RBDC}&\multirow{2}{*}{TBDC}&\multicolumn{2}{c}{AUC}&\multirow{2}{*}{RBDC}&\multirow{2}{*}{TBDC}\\
			 \cmidrule(lr){3-4}\cmidrule(lr){7-8}\cmidrule(lr){11-12}
    		 && \small Micro &\small Macro &&&\small Micro &\small Macro &&&\small Micro &\small Macro &&\\
    		 \midrule
    		 
    		 \multirow{6}{*}{\rotatebox[origin=c]{90}{\small Frame / Patch level}}&Liu \textit{\textit{et al.}} \cite{future_pred} (pred.)&95.4&-&-&-&72.8&-&-&-&85.1&-&-&-\\
    		 &Nguyen \textit{\textit{et al.}} \cite{app-mo-co} (pred. \& rec.)&96.2&-&-&-&-&-&-&-&86.9&-&-&-\\
    		 &Gong \textit{\textit{et al.}} \cite{memae} (rec.)&94.1&-&-&-&71.2&-&-&-&83.3&-&-&-\\
    		 &Park \textit{\textit{et al.}} \cite{mnad} (pred. \& rec.)&-&97.0&-&-&-&72.0&-&-&-&88.5&-&-\\
    		 &Ramachandra \textit{\textit{et al.}} \cite{streetscene} (dist.)&\multicolumn{2}{c}{88.3}&62.5&80.5&-&-&-&-&\multicolumn{2}{c}{72.0}&35.8&\textbf{80.9}\\
    		 
    		 &Ramachandra \textit{\textit{et al.}} \cite{seamese} (dist.) &\multicolumn{2}{c}{94.0}&\underline{74.0}&80.3&-&-&-&-&\multicolumn{2}{c}{87.2}&41.2&\underline{78.6}\\
    		 \midrule
      		 \multirow{7}{*}{\rotatebox[origin=c]{90}{\small Object level}}&Ionescu \textit{\textit{et al.}} \cite{dummyae} (rec.)&94.3&97.8&52.76&72.88&78.7&84.9&20.7&44.5&87.4&90.4&15.8&27.0\\

    		  &Yu \textit{\textit{et al.}} \cite{cloze_test} (pred.)&97.3&-&-&-&74.8&-&-&-&89.6&-&-&-\\
    		 
    		 &Liu \textit{\textit{et al.}} \cite{hybrid}(pred. \& rec.)&\multicolumn{2}{c}{99.3}&-&-&\multicolumn{2}{c}{76.2}&-&-&\multicolumn{2}{c}{91.1}&-&-\\
    		
    		 &Georgescu \textit{\textit{et al.}} \cite{baf} (rec.) &\underline{98.7}&99.7&69.2&\underline{93.2}&\textbf{82.7}&\textbf{89.3}&41.3&78.8&92.3&90.4&\underline{65.1}&66.9\\
    		 &Georgescu \textit{\textit{et al.}} \cite{ssmt} (multi.)&97.5&\underline{99.8}&72.8&91.2&\underline{82.4}&\textbf{89.3}&\underline{42.8}&\underline{83.9}&91.5&91.9&57.0&58.3\\
    		 \cmidrule(lr){2-14}
    		 &Ours &\textbf{98.9}&\textbf{99.9}&\textbf{77.2}&\textbf{98.5}&77.1&\underline{86.2}&\textbf{51.6}&\textbf{84.6}&92.6*&90.1*&\textbf{75.3}&73.4\\
    		\bottomrule
			
		\end{tabular}%
    }
    \vspace{1mm}
    \caption{Comparison of our approach with the state-of-the-art methods (\%) on the metrics: Micro AUC, Macro AUC, RBDC and TBDC. The AUC scores for the methods which do not explicitly mention the type (Micro or Macro) are put in the middle of the column. Best results are in bold, second best are underlined. Methods are grouped according to whether they detect anomalies at the frame/patch or object level. We also classify approaches according to the used paradigm (pred.: prediction based, rec.: reconstruction based, dist.: distance based, and multi.: multiple pretext tasks). 
    * The ground truth provided by Ramachandra et al. \cite{streetscene} is used for computing the AUC for Avenue as mentioned in section 4.1 }
	\label{tab:results}

\end{table*}

\subsection{Evaluation}

In order to evaluate the frame level performance of our method, we adopt the area under the receiver characteristic curve (AUC ROC) which is a popular metric in the VAD literature. As pointed out in \cite{baf}, many previous works do not specify if the AUC is computed at the video level and averaged across all the dataset (Macro AUC) or at the dataset level by concatenating all video frames (Micro AUC). We report both measures for clarity. 

The AUC quantifies the anomaly detection performance at the frame level only and not the ability of the model to localize and track anomalies. In fact, if the model predicts a false positive in an abnormal frame, it will be counted as a true positive. Previous works such as \cite{ganicip17} reported the pixel level AUC as an attempt to measure the anomaly localization performances. This metric considers a detection as a true positive if at least 40\% of the ground truth pixels are detected. As pointed out in \cite{streetscene}, this procedure is problematic since it takes into account only false positive detections in normal frames and not in abnormal ones. As an alternative to pixel level AUC, we adopt the region-based detection criterion (RBDC) introduced by Ramachandra \textit{\textit{et al.}} \cite{streetscene} since it takes into account false positives detections in all frames. We report also the track-based detection criterion (TBDC) \cite{streetscene} which quantifies the models ability to identify anomalies spatio-temporally. Since, the previous metrics evaluate the ability to localize and track anomalies, we consider them as the main evaluation criteria for abnormal event detection.   

\subsection{Implementation details}

For the pre-processing step, we choose YOLOv3 \cite{yolov3} object detector as in \cite{ssmt,baf,dummyae} \blue{for a fair comparison with them.} We use a MMDetection \cite{mmdetection} implementation of YOLOv3 pretrained on MS COCO dataset. We set the confidence threshold to 0.7 for Avenue and ShanghaiTech. Since objects have a low resolution in UCSDped2 we decreased the confidence threshold to 0.5 as in \cite{baf}. For this dataset, we notice that the implementation of YOLOv3 produced very small false positive boxes, we filtered them out using an area criterion (300 pixels). Optical flow between consecutive frames is extracted using an OpenCV implementation of Gunnar Farneback's algorithm \cite{of}, which offers a good trade-off between accuracy, speed and generalisation across multiple contexts. \green{In order to normalize the optical flow magnitudes between 0 and 1, we divide them by 50 which is an upper-bound over the maximum displacement in the training videos}. For all datasets, we choose a time step $\delta t = 1$ for past and future prediction pretext tasks. As in \cite{ssmt,baf,dummyae}, we resize objects to 64x64.

Regarding our architecture the encoder $\mathcal{E}$ consists of five 3x3 convolution layers followed by 2x2 max pooling and ReLU activation. We double the number of convolution filters after each layer so that we ensure that the architecture is not constrained to perform dimensionality reduction and therefore our approach is different from the reconstruction-based methods. The four decoding branches are identical except for the last layer that matches the output number of channels. They are composed of four transposed convolutions with a kernel size of 4x4 and a stride of 2 followed by ReLU activations.

For all datasets, we trained the network for 200 epochs using Adam optimizer \cite{adam} with a learning rate of $10^{-3}$, a batch size of 640. The frame level scores are smoothed using a Gaussian filter as in \cite{ssmt,baf}. RBDC and TBDC are computed using the code provided by Georgescu \textit{\textit{et al.}} \cite{baf}. It is worth mentioning that our anomaly detection approach achieves state-of-the-art performances with few hyperparameters.

%------------------------------------------------------------------------ 
\subsection{Results}\label{sec:res}

Table \ref{tab:results} reports quantitative evaluations of our method on UCSDped2 \cite{ped}, ShanghaiTech \cite{shanghaitech} and Avenue \cite{avenue} benchmarks, in comparison with state-of-the-art methods \cite{ssmt,baf,hybrid,cloze_test,dummyae,seamese,streetscene,memae,app-mo-co,future_pred}. Methods are grouped according to the used paradigm and whether they perform anomaly detection at the object or frame/patch level. It is important to mention that no method consistently outperforms the others in all metrics and in all datasets.

\begin{figure}[h]
    \centering
    \begin{minipage}[b]{0.48\textwidth}
        \center \includegraphics[width=0.8\textwidth]{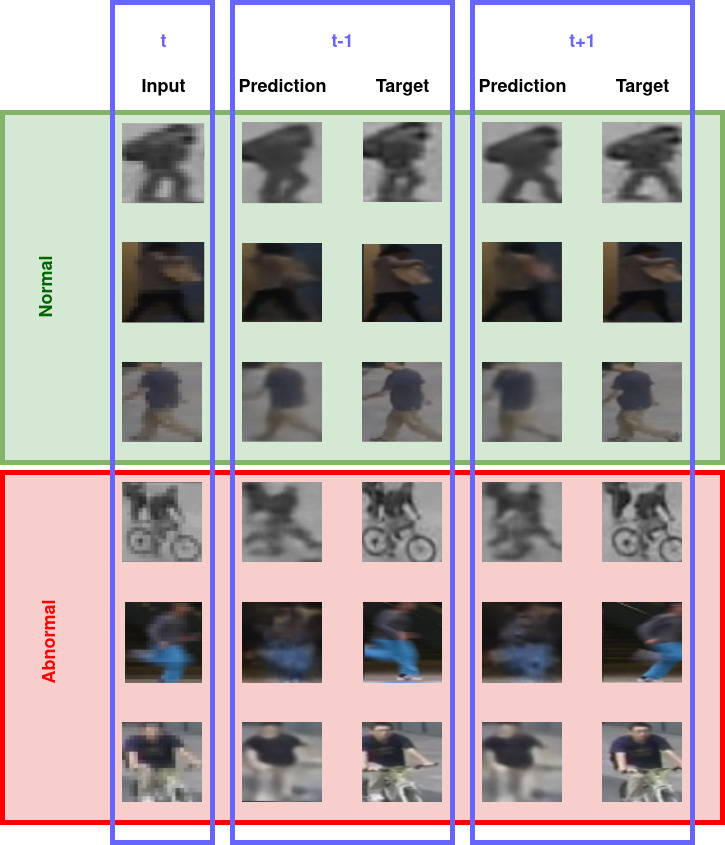}
        \vspace{2mm}
        \caption{Normal and abnormal test samples with their corresponding predictions. Testing examples are taken from the datasets UCSDped2 \cite{ped}, Avenue \cite{avenue} and ShanghaiTech \cite{shanghaitech}. The first column indicates the down-scaled inputs at time $t$. The second column shows the past frame predictions and the corresponding ground truth at time $t-1$. The last column shows the future frame predictions with the ground truth $t+1$. Appearances and behaviours are well predicted for normal samples while it is not the case for anomalies.}
        \label{fig:preds}

    \end{minipage}
\end{figure}

Qualitative results are provided in figure \ref{fig:preds}. We can observe that appearances and behaviours are well predicted for normal samples while it is not the case for anomalies, which allows to distinguish them.

\paragraph*{Results on UCSDped2}

\begin{figure}[h]
    \centering
    \begin{minipage}[b]{0.45\textwidth}
        \includegraphics[width=1\textwidth,trim={2.5cm 0cm 2.5cm 0cm},clip]{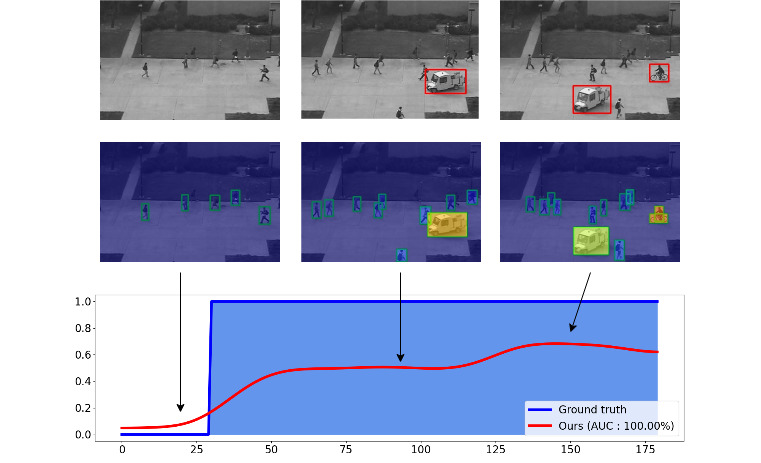}
        \vspace{2mm}
        \caption{Qualitative results from the video Test004 taken from UCSDped2. Anomalies consist of a car and a cyclist introduced in a pedestrian area. Red boxes indicate the ground truth and green boxes indicate object detections}
        \label{fig:ped2_auc}
    \end{minipage}
\end{figure}

Recent works reported AUC performances which are higher than 98\% on this dataset, while, the metrics RBDC and TBDC show a better discrimination between approaches in terms of anomaly localization and tracking than the AUC. As shown in table \ref{tab:results}, our approach is able to significantly outperform previous approaches on anomaly tracking by 3 p.p on the RBDC metric and 5 p.p on TBDC. Those improvements can be explained by the choice of pretext tasks which constrain further the anomaly prediction, at the same time, our pretext tasks leverage both motion and appearance which are relevant for anomalies present in this dataset. We observed that the optical flow prediction error is particularly relevant for this dataset since it achieves a Macro AUC of 99.8\% as shown in the ablative study in section \ref{sec:ablative}. Adding the appearance information allows the model to achieve state-of-the-art performances on all metrics. The figure \ref{fig:ped2_auc} illustrates our model performances. 

\begin{figure}[h]
    \centering
    
    \begin{minipage}[b]{0.45\textwidth}
        \includegraphics[width=1\textwidth,trim={2.8cm 0cm 2.8cm 0cm},clip]{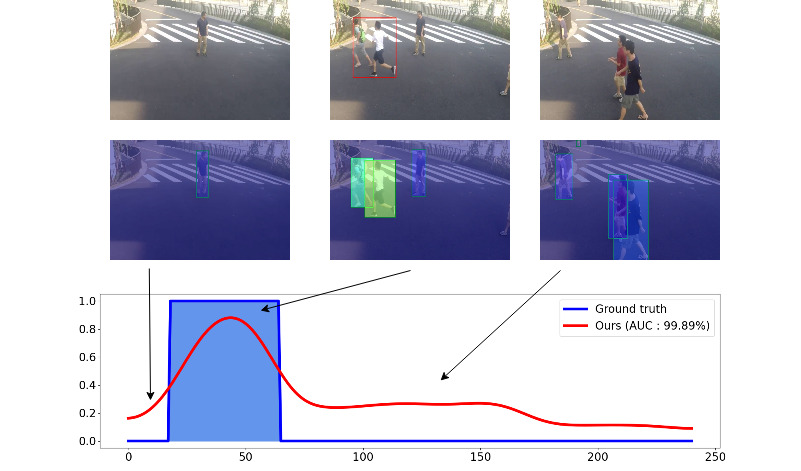}
        \caption{Qualitative results from the video 03\_0061 taken from ShanghaiTech. The anomaly consists in one person running after another. The red box indicates the ground truth and green boxes indicate object detections}
        \label{fig:stc_auc}
    \end{minipage}
\end{figure}

\begin{figure}[b]
    
    \centering
    \begin{minipage}[b]{0.44\textwidth}
        \includegraphics[width=1\textwidth,trim={2.5cm 0cm 3cm 0cm},clip]{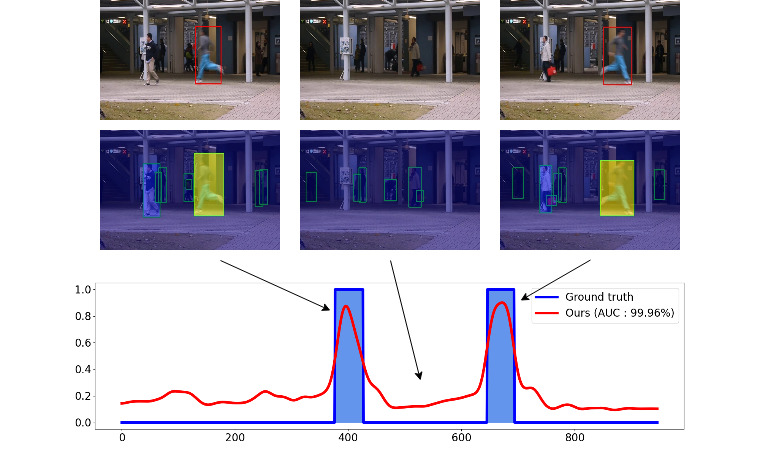}
        \caption{Qualitative results from the video 04 taken from Avenue, the anomaly is represented by a running person. Red boxes indicate the ground truth and green boxes indicate object detections}
        \label{fig:avenue_auc}
    \end{minipage}
\end{figure}

\begin{table*}[t]
        \centering
        \resizebox{1.8\columnwidth}{!}{%
		\begin{tabular}{ccccccccc}
			\toprule
			Ablative&\multicolumn{2}{c}{pretext tasks} & \multicolumn{3}{c}{Down-scaling ratios} & \multicolumn{3}{c}{Evaluation metrics} \\
			\cmidrule(lr){2-9}Variant&{Appearance} & {Flow Mag.} & {$\frac{1}{2}$} &{$\frac{1}{4}$} & {$1$}
			&Micro AUC&RBDC&TBDC\\
			\midrule
			Best model & \checkmark & \checkmark& \checkmark & & &\textbf{98.9}&\textbf{77.2}&\textbf{98.5}\\

			& \checkmark & \checkmark &  & \checkmark& &98.5&77.1&96.3\\
			w/o down-scaling & \checkmark & \checkmark & & & \checkmark&98.1&73.3&95.0\\
			\midrule
			w/o Flow Mag.& \checkmark & \ding{55} & \checkmark&& &95.0&64.4&77.2\\
			w/o Appearance & \ding{55}  & \checkmark & \checkmark  & & &96.9&73.1&{98.0}\\
			\bottomrule
		\end{tabular}%
		}
		\vspace{1mm}
		\caption{Micro AUC, RBDC, TBDC scores in (\%) obtained on UCSDped2 by removing various component of our method. Best performances are in bold.}
		\label{tab:ablative}
\end{table*}

\paragraph*{Results on ShanghaiTech} This is  one of the most challenging datasets due to its diversity of anomalies and scene changing. Our method achieves significant improvement of 8 p.p in terms of RBDC and  slightly outperforms other methods on TBDC. We notice that our method surpasses approaches that use the future frame prediction pretext task such as \cite{future_pred,hybrid,cloze_test} in terms of AUC which shows the relevance of further constraining those tasks by discarding past frames and performing down-scaling. In addition, we observe that approaches surpassing our method in terms of AUC \cite{ssmt,baf} introduce an additional supervision such as pseudo anomalies in the case of \cite{baf} or model distillation \cite{ssmt} where the anomaly detection model is trained on YOLO class prediction. Our approach requires less supervision since 1) object detection is performed only for localisation and not for classification, 2) no pseudo-abnormal data are used. Figure \ref{fig:stc_auc} shows an example of anomaly localization and tracking on this dataset. Our scores follow well the temporal and spatial ground truth annotations.

\paragraph*{Results on Avenue} \label{sec:avenue_res} In terms of localization, we achieve a significant improvement of 13 p.p on RBDC over the state-of-the-art, while being close in terms of TBDC. The difference in tracking performances between the current state-of-the-art \cite{streetscene,seamese} and our method can be explained by the fact that the temporal context used in \cite{streetscene,seamese} is larger than ours which leads to a better tracking performances. Nevertheless, compared to object centric approaches such as \cite{baf,ssmt}, which uses an equivalent temporal context, we improve the TBDC by 6 p.p. We explain those improvements by two main reasons. First, we found that down-scaling the input considerably enhances RBDC by 10 p.p. Second, we noticed that adding the scale information to the anomaly scoring is beneficial for this dataset since anomalies in Avenue depend on their location with respect to the camera \green{pose (scene layout) unlike in the UCSDped2 and Shanghaitech datasets for which the location information is less relevant to detect anomalies}. Therefore, we take into account scale change by multiplying \green{object level} anomaly scores \green{$S$} by the bounding boxes width, which leads to a substantial improvement of 12 p.p on RBDC. The figure \ref{fig:avenue_auc} shows an illustration of our model performances.

\section{Ablation study} \label{sec:ablative}

We performed an ablation study to evaluate the impact of each component of our method (Table \ref{tab:ablative}). First, we tested the impact of removing the pretext tasks related to appearance and motion respectively. We notice that in both cases, the performances decrease, showing the complementary roles of appearance and motion in detecting abnormal events. In order to assess the impact of input down-scaling, \blue{ we tried two down-scaling ratios : $\frac{1}{2}$ and $\frac{1}{4}$. We observe that the ratio $\frac{1}{2}$ gives slightly better performances. One possible explanation is that it offers a good compromise in terms of predictability. In fact, too much down-scaling makes the prediction hard even for normal samples, while without down-scaling it is easier for the model to recover anomalies.} Indeed, we observe a global drop in performances which is particularly high for RBDC (4.9p.p) when no down-scaling is performed. This results show that constraining further the prediction task by down-scaling the input is useful.

\section{Conclusions}

In this work, we introduced a new way of approaching the VAD problem, by imposing constrained pretext tasks to learn appearance and motion normality patterns. The experiments conducted on benchmark datasets show the effectiveness of our methodology. In future work, we will explore more pretext tasks in order to further improve VAD performances and address new types of anomalies. 

{\small
\bibliographystyle{ieee}
\bibliography{egbib}
}

\end{document}